\documentclass{article}

\usepackage{arxiv}
\date{}
\renewcommand{\today}{}

\usepackage[utf8]{inputenc} 
\usepackage[T1]{fontenc}    
\usepackage{hyperref}       
\usepackage{url}            
\usepackage{booktabs}       
\usepackage{amsfonts}       
\usepackage{nicefrac}       
\usepackage{microtype}      
\usepackage{amsmath}        
\usepackage{lipsum}
\usepackage{graphicx}
\graphicspath{ {./images/} }
\usepackage{float}
\usepackage{natbib}
\usepackage{xcolor}
\usepackage{wrapfig}
\definecolor{Highlight}{RGB}{0,150,0}

\usepackage{algorithm}
\usepackage{algpseudocode}   
\usepackage{caption}   
\setcitestyle{numbers}

\definecolor{hotpurple}{RGB}{180, 80, 200}
\newcommand\ourmethod{CVG}

\title{Compositional Video Generation via Inference-Time Guidance}

\author{
  Ariel Shaulov$^{1}$\thanks{Equal contribution, Corresponding author: arielshaulov@mail.tau.ac.il} \quad
  Eitan Shaar$^{2*}$ \quad
  Amit Edenzon$^{3*}$ \quad
  Gal Chechik$^{3,4}$ \quad
  Lior Wolf$^1$ \\ \\
$^{1}$Tel-Aviv University \quad
$^{2}$Independent Researcher \quad
$^{3}$Bar Ilan University  \quad
$^{4}$NVIDIA Research \quad
}

\begin{document}
\maketitle
\vspace{-30pt}
\begin{center}
{\large \href{https://arielshaulov.github.io/CVG/}{\textcolor{hotpurple}{\textbf{CVG Project Page}}}}
\end{center}

\begin{figure}[H]
  \includegraphics[width=\textwidth]{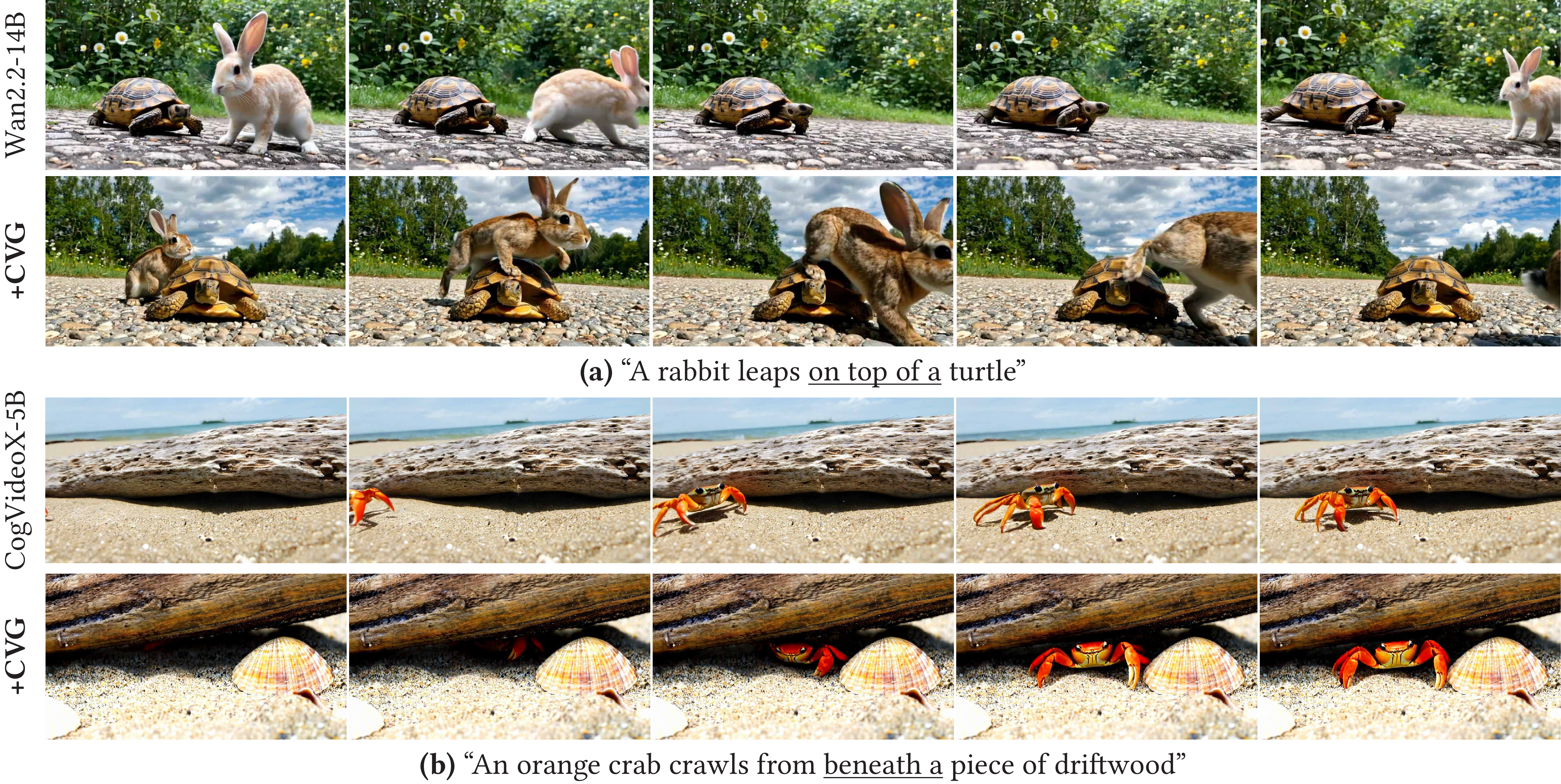}
  \caption{Text-to-video results before and after applying CVG on Wan2.2-14B \cite{wan2025wan} and CogVideoX-5B \cite{yang2024cogvideox}}
  \label{fig:teaser}
\end{figure}

\begin{abstract}
Text-to-video diffusion models generate realistic videos, but often fail on prompts requiring fine-grained compositional understanding, such as relations between entities, attributes, actions, and motion directions. We hypothesize that these failures need not be addressed by retraining the generator, but can instead be mitigated by steering the denoising process using the model's own internal grounding signals. We propose \textbf{\ourmethod{}}, an inference-time guidance method for improving compositional faithfulness in frozen text-to-video models. Our key observation is that cross-attention maps already encode how prompt concepts are grounded across space and time. We train a lightweight compositional classifier on these attention features and use its gradients during early denoising steps to steer the latent trajectory toward the desired composition. Built on a frozen VLM backbone, the classifier transfers across semantically related composition labels rather than relying only on narrow category-specific features. \ourmethod{} improves compositional generation without modifying the model architecture, fine-tuning the generator, or requiring layouts, boxes, or other user-supplied controls. Experiments on compositional text-to-video benchmarks show improved prompt faithfulness while preserving the visual quality of the underlying generator.

\end{abstract}

\section{Introduction}

Text-to-video (T2V) diffusion
models~\cite{ho2020ddpm,rombach2022ldm,ho2022video,yang2024cogvideox,
wang2025wan,guo2023animatediff} have achieved substantial progress
in visual fidelity and temporal coherence. Their performance,
however, degrades on prompts that require fine-grained
\emph{compositional} understanding, where multiple entities,
attributes, actions, and interactions must be realized correctly in
a single video. Given prompts such as ``a dog turning left'' or ``a
cat walking away from a car,'' state-of-the-art generators often
produce videos that are visually plausible but compositionally
incorrect: the dog turns in the wrong direction, or the cat moves
toward rather than away from the car. This limitation is
consistently reflected in recent compositional T2V
benchmarks~\cite{sun2025t2v,huang2025vchain,shaulov2025flowmo}, and
constitutes a central failure mode of current video generators.

Compositional prompts are challenging because they require
attributes, motion, spatial configuration, and interactions to be
realized jointly over time. In practice, standard text conditioning
and classifier-free guidance provide a global prompt-level signal, but
do not explicitly verify whether each fine-grained compositional
constraint is satisfied. This has motivated a growing line of work on
stronger controllability, either by retraining the generator to accept
auxiliary control signals such as bounding boxes, layouts, or motion
trajectories~\cite{li2023gligen,wang2024boximator,
yin2023dragnuwa,geng2025motionprompting}, by aligning it to reward
models through fine-tuning~\cite{black2023ddpo,clark2024draft,
prabhudesai2024vader}, or by scaling inference-time
search~\cite{ma2025inftscale,liu2025video,he2025scaling}. These
strategies can improve compositional control, but often require
additional training at video scale, new annotated data, user-supplied
conditioning, or expensive sampling-time compute.

We propose an alternative that requires none of these. Our approach
is motivated by the observation that cross-attention maps associated
with the relevant prompt tokens already encode much of the
spatiotemporal information needed to evaluate compositional
correctness. The cross-attention of tokens such as \textit{dog},
\textit{car}, or action-related tokens indicates, frame by frame,
where the model localizes the corresponding concepts, and how these
localizations evolve over
time~\cite{hertz2022prompt,chefer2023attendandexcite}. This suggests
that a lightweight classifier trained on these attention maps can
serve as a differentiable predictor of compositional correctness,
whose gradient can then be used to steer the denoising trajectory at
inference time.

Building on this observation, we introduce \ourmethod{}
(\textbf{C}ompositional \textbf{V}ideo \textbf{G}eneration), an
inference-time control framework for compositional T2V diffusion.
\ourmethod{} leaves the pretrained generator frozen and trains a
compact VLM-based compositional classifier over cross-attention
features extracted from real and synthetic
videos~\cite{shang2019vidor,shang2017vidvrd,huang2025vchain}.Since
the classifier is built on top of a frozen VLM backbone, it inherits
a broad vision-language prior that helps it generalize to
semantically related composition labels beyond the exact compositional labels seen
during training. Thus, at sampling time, the classifier can be used as a differentiable controller: at each guided denoising
step, a cross-entropy loss between the classifier's prediction and
the target compositional label is backpropagated to update the
latent. To prevent the classifier from exploiting superficial
textual traces in the attention signal, an effect we term
\emph{composition leakage}, we adopt a video analogue of dual
inversion~\cite{yiflach2025data} during training. Guidance is
restricted to the early denoising steps, during which coarse motion
and scene structure are established, thereby preserving the
appearance priors of the underlying generator.



In this paper, we make the following contributions. 
(i) We propose \ourmethod{}, an inference-time control framework for compositional T2V diffusion that steers a frozen generator using gradients from a lightweight compositional classifier, without fine-tuning, auxiliary conditioning, layout input, or user-provided spatial controls. 
(ii) We introduce a VLM-based compositional classifier trained on subject-token cross-attention maps from real and synthetic videos. We obtain real-video attention maps through video inversion and use dual inversion to reduce composition leakage. By leveraging the frozen VLM prior, the classifier generalizes to semantically related composition labels beyond those seen during training.
(iii) On compositional T2V benchmarks~\cite{sun2025t2v}, \ourmethod{} consistently improves compositional faithfulness over both the base generators and the layout-guided TTOM~\cite{qu2025ttom} baseline, while preserving the visual quality of the underlying generator.

\section{Related Work}

\paragraph{\textbf{Text-to-Video Generation.}}
Modern text-to-video (T2V) generation is dominated by diffusion-based
and autoregressive
paradigms~\cite{ho2022video,ho2022imagen,singer2022make,
wang2025lavie,gupta2024photorealistic,
yang2024cogvideox,jin2024pyramidal,wang2025wan,kondratyuk2023videopoet,
yin2025slow,guo2023animatediff}. Early diffusion-based systems such
as Video Diffusion Models~\cite{ho2022video}, Imagen
Video~\cite{ho2022imagen}, and
Make-A-Video~\cite{singer2022make} established the basic recipe
of jointly modeling space and time via diffusion cascades with
super-resolution stages. Subsequent latent-diffusion and
transformer-based systems, including
LaVie~\cite{wang2025lavie},
CogVideoX~\cite{yang2024cogvideox}, and
Wan~\cite{wang2025wan}, improved fidelity, motion realism, and
prompt alignment through stronger temporal modeling and large-scale
pretraining. Autoregressive and causal generators produce frames or
tokens sequentially, making them particularly suited for streaming
and long-context
synthesis~\cite{kondratyuk2023videopoet,yin2025slow,huang2025self,
wang2024loong,lin2024open}. Our method is agnostic to the specific
backbone: it treats the pretrained T2V generator as a black box and
introduces no modifications to its weights or sampling schedule
beyond an additive guidance term, making it applicable across
diffusion and flow matching based
models~\cite{yang2024cogvideox,wang2025wan}.

\paragraph{\textbf{Compositional Visual Generation.}}
Compositional generation has been studied extensively in the image
domain, beginning with energy-based and compositional diffusion
formulations that combine multiple concepts and
relations~\cite{du2020compositional,liu2022compositional}, and
continuing with training-free attention-control methods that improve
attribute binding and subject
localization~\cite{hertz2022prompt,chefer2023attendandexcite,
he2023localized}. Structural conditioning approaches such as
ControlNet~\cite{zhang2023controlnet},
GLIGEN~\cite{li2023gligen}, LayoutDiffusion~\cite{zheng2023layoutdiffusion},
and MultiDiffusion~\cite{bartal2023multidiffusion} instead train
auxiliary modules to accept bounding boxes, layouts, or segmentation
masks as additional inputs. Extending compositionality to video is
substantially more challenging, as the model must bind multiple
entities and relations while preserving temporal coherence and
aligning events with the correct time spans. Recent work has
therefore explored compositional video synthesis with spatial and
temporal conditioning~\cite{wang2023videocomposer,wang2024mvoc},
layout planning for multi-object
scenes~\cite{tian2024videotetris,he2025dyst,yang2024compositional},
and temporally grounded multi-event
generation~\cite{oh2024mevg,wu2025mind}. Nevertheless, recent
benchmarks indicate that current T2V models continue to struggle
with attribute binding, motion binding, spatial relations, and
object interactions~\cite{sun2025t2v,huang2025vchain}. Our work addresses this broader compositional generation problem, improving the faithful realization of complex prompts without auxiliary conditioning inputs or retraining the generator.


\paragraph{\textbf{Inference-Time Guidance for Diffusion Models.}}
Inference-time guidance offers a training-free route to shaping
generation by intervening in the sampling process itself. In the
image domain, classifier and classifier-free
guidance~\cite{dhariwal2021diffusionbeatgans,ho2022classifierfree}
established the basic template of modifying score estimates with
gradient signals, and subsequent work extended this framework in
several directions: cross-attention control for structural
editing~\cite{hertz2022prompt,he2023localized}, attention-based
semantic nursing~\cite{chefer2023attendandexcite}, direct latent
optimization~\cite{wallace2023doodl,shi2024dragdiffusion}, universal
guidance through off-the-shelf classifiers and energy
functions~\cite{bansal2023universal,yu2023freedom}, score
distillation~\cite{poole2023dreamfusion}, and dynamic guidance
scheduling~\cite{shtedritski2025dynamiccfg}. Concurrent efforts on
inference-time scaling show that search over noise candidates or
denoising trajectories can yield substantial quality improvements
without additional
training~\cite{ma2025inftscale,zhuo2025reflectionflow}. These ideas
have begun to migrate to video generation, through variance-based
flow guidance~\cite{shaulov2025flowmo}, beam search and test-time
scaling~\cite{oshima2025inference,liu2025video,he2025scaling,
fang2025inflvg}, and test-time
optimization~\cite{qu2025ttom,yiflach2025data}. These ideas
have begun to migrate to video generation, through variance-based
flow guidance~\cite{shaulov2025flowmo}, beam search and test-time
scaling~\cite{oshima2025inference,liu2025video,he2025scaling,
fang2025inflvg}, and test-time
optimization~\cite{qu2025ttom,yiflach2025data}. Other
inference-time methods instead target temporal consistency,
long-horizon coherence, or sampling efficiency through
noise-initialization, frequency-domain, and caching
strategies~\cite{wu2024freeinit,lu2024freelong,ren2024consisti2v,
liu2024teacache,fan2025taocache,shaulov2026tokentrim}. Our work is also a inference-time guidance method for video generation, and in particular to approaches that improve compositional fidelity without retraining.

\begin{figure*}[t]
  \centering
  \includegraphics[width=\textwidth]{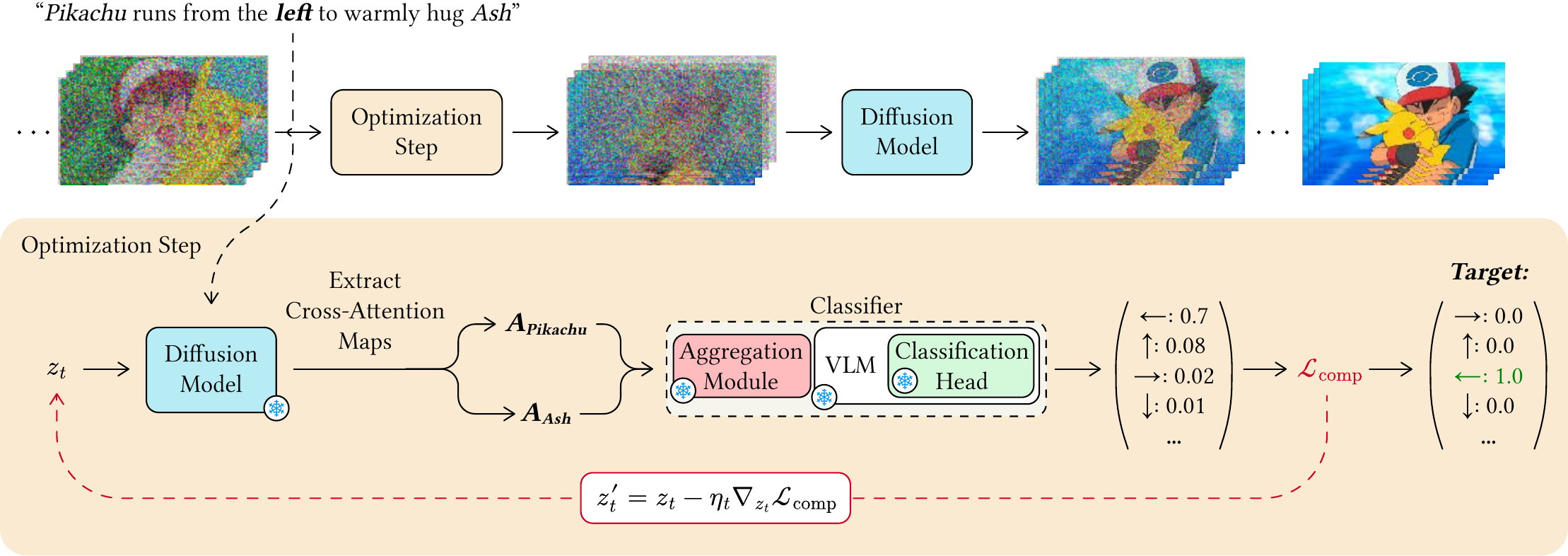}
    \caption{
    \textbf{Overview of \ourmethod{}.}
    At inference time, \ourmethod{} extracts subject-token cross-attention maps from a frozen text-to-video diffusion model and feeds them to a lightweight composition classifier. The classifier predicts the current compositional relation, and its loss with respect to the target compositional relation is backpropagated to update the latent. This provides composition-aware guidance without fine-tuning the generator or requiring external layout controls.
  }
  \label{fig:method_figure}
\end{figure*}

\begin{figure*}[t]
  \centering
  \includegraphics[width=\textwidth]{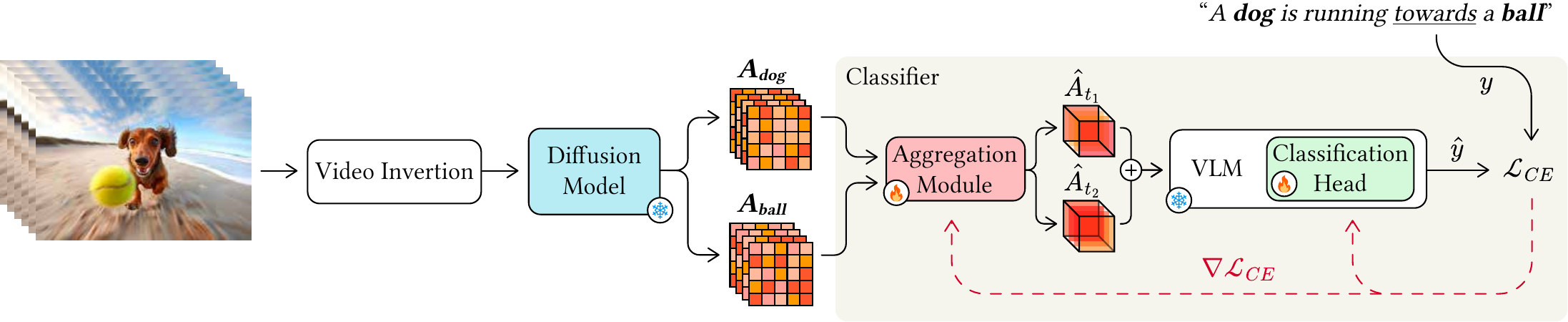}
  \caption{
     \textbf{Training compositional classifier.}
    Given a real video and its prompt, we invert the video into the latent space of the frozen text-to-video model and extract subject-token cross-attention maps. These maps are processed by an aggregation module, a frozen VLM, and a trainable classification head to predict the target compositional relation label using cross-entropy supervision.
  }
  \label{fig:training_figure}
\end{figure*}

\section{Method}
\label{sec:method}

We present \textbf{\ourmethod{}}, an inference-time control framework for text-to-video generation. \ourmethod{} is based on the observation that pretrained text-to-video diffusion models expose useful grounding signals through their internal cross-attention maps: these maps indicate how prompt entities are localized and how their spatial support evolves during denoising. Rather than fine-tuning the generator or introducing external layout controls, we train a lightweight compositional classifier on top of these attention maps, and use its gradient at inference time to guide the denoising process toward a desired target composition. 

A key design choice in \ourmethod{} is that the classifier receives only the cross-attention maps of the \emph{composition-relevant entity tokens}. For example, for the prompt ``a dog moves behind a ball,'' the classifier is given the cross-attention maps corresponding to both \textit{dog} and \textit{ball}, and is trained to predict the ground-truth label \textit{behind}. Similarly, for prompts involving multiple entities, such as ``Pikachu stands next to Ash,'' the classifier receives the attention maps of both entities rather than the relation token itself. In this way, the classifier learns to infer compositions from the relative spatiotemporal evolution of the entities' grounding, rather than by directly reading the target composition token from the prompt. During inference, the same classifier is applied to the evolving latent video, producing a score over all composition classes, and its cross-entropy loss with respect to the desired target is backpropagated to update the current latent. See method overview in Fig~\ref{fig:method_figure}.

\subsection{Problem Formulation}
\label{sec:problem_formulation}

Let $p$ denote a text prompt describing a subject and a target relation. For example, $p$ may be ``a dog turning left.'' Given a pretrained text-to-video diffusion model $G$, our goal is to generate a video $x$ that is both visually plausible and semantically aligned with the target composition specified in the prompt.

Let $y \in \mathcal{Y}$ denote the target composition label extracted from the prompt, where $\mathcal{Y}$ is an open vocabulary of composition classes. In our setting, these labels may encode directional or spatial semantics such as \textit{left}, \textit{right}, \textit{above}, or \textit{below}, see full list of compositions in App.~\ref{app:Training_Composition_Labels}. Standard text conditioning alone often does not guarantee accurate control over such fine-grained semantics. Our objective is therefore to guide the denoising trajectory such that the generated video is recognized by a learned classifier as expressing the target composition $y$.

Formally, at a selected denoising step $t$, let $z_t$ denote the current latent video variable. \ourmethod{} defines a learned composition loss $\mathcal{L}_{\mathrm{comp}}(z_t,p,y)$, and updates the latent by backpropagating this loss:
\begin{equation}
z_t' = z_t - \eta_t \nabla_{z_t} \mathcal{L}_{\mathrm{comp}}(z_t,p,y),
\label{eq:latent_guidance_update}
\end{equation}
where $\eta_t$ is the guidance step size. The role of the method is therefore to construct a compositional classifier whose output can serve as an effective differentiable objective for inference-time control.

\subsection{\ourmethod{}}
\label{sec:\ourmethod{}}

\ourmethod{} consists of two components:
\begin{enumerate}
    \item \textbf{Frozen text-to-video generator.} We use a pretrained text-to-video diffusion model as a generative prior over appearance, motion, and temporal dynamics. Its parameters remain fixed throughout training and inference.
    \item \textbf{Compositional classifier.} We train a classifier whose input is the subject-token cross-attention extracted from the frozen generator. The classifier is composed of a learned aggregation module, a frozen video-language-model (VLM) backbone, and a trainable classification head.
\end{enumerate}

The classifier is trained in a supervised manner to predict composition labels from subject-token cross-attention maps. Importantly, the text-to-video generator is never optimized during classifier training. The generator only serves as a source of structured internal representations.

More concretely, for a prompt $p$, we first extract the subject-token cross-attention maps from a selected set of denoising steps. These maps are then processed by a learned aggregation module that compresses them into a compact spatiotemporal representation. This representation is fed into a frozen VLM backbone, followed by a trainable classification head that outputs a distribution over compositional classes.

During training, only the aggregation module and the classification head are optimized, while both the generator and the VLM backbone remain frozen. During inference, the same classifier is reused as a differentiable controller: at selected denoising steps, we extract the relevant subject-token cross-attention maps, obtain the classifier's scores over all compositional classes, and optimize the latent using a cross-entropy loss toward the target composition. See training overview in Fig~\ref{fig:training_figure}.

\subsection{Cross-Attention as a Composition Representation}
A central design choice in \ourmethod{} is to learn compositional control from the internal cross-attention maps of a frozen text-to-video diffusion model. These maps capture how prompt entities are grounded in the evolving spatiotemporal latent, making them a natural representation for inferring spatial and motion-based compositions. For each selected denoising step $t \in \mathcal{T}$, layer $\ell \in \mathcal{L}$, head $h \in \mathcal{H}$, and composition-relevant entity token $k \in \mathcal{K}(p)$, we extract cross-attention maps
\begin{equation}
A_{t,\ell,h}^{(k)} \in \mathbb{R}^{F \times H \times W},
\end{equation}
where $F$ is the number of frames and $H \times W$ is the latent spatial resolution. Importantly, the classifier receives only the entity-token maps, not the composition token itself: for ``a dog moves behind a ball,'' it receives the maps of \textit{dog} and \textit{ball}, while the label \textit{behind} is withheld. For prompts involving more than two entities, such as ``a dog stands between a ball and a bench,'' the classifier receives the maps of all composition-relevant entities, e.g., \textit{dog}, \textit{ball}, and \textit{bench}. The extracted maps are aggregated by a learned module into a compact representation $\phi(z_t,p)$, which is passed through a frozen VLM backbone and a trainable classification head to predict the target composition. We train the classifier on both real and synthetic videos. For real videos, we obtain the required cross-attention maps by first applying video inversion and then replaying the denoising trajectory, allowing the classifier to learn from natural motion patterns and object interactions. For synthetic videos, we use generated examples with controlled composition labels, which increase coverage over rare spatial relations, motion directions, and multi-object configurations. Together, the real and synthetic data provide both realistic dynamics and diverse compositional supervision. Since the classifier is built on top of a frozen VLM backbone, it also inherits a strong semantic prior from large-scale vision-language pretraining, enabling useful generalization beyond the exact composition labels observed during training. To prevent \emph{composition leakage}, where the classifier predicts the relation from textual traces in the attention maps rather than from the visual spatiotemporal arrangement of the grounded entities, we adopt a video analogue of dual inversion~\cite{yiflach2025data}. Specifically, each training video is inverted once with the correct composition prompt and once with an incorrect composition prompt sampled from the label vocabulary, while both inversions are supervised with the same ground-truth composition label. This forces the classifier to rely on the actual visual grounding of the entities rather than the textual relation word. Unlike~\cite{yiflach2025data}, which uses dual inversion for image-level inference-time optimization, we use it to train a video compositional classifier over spatiotemporal cross-attention maps extracted from real videos. Full details described in App.~\ref{Additional_Implementation_Details_For_Compositional_classifier}.

Since the classifier is built on top of a frozen VLM backbone, it inherits a strong semantic prior from large-scale vision-language pretraining. This allows it to generalize beyond the exact composition labels observed during training, providing useful guidance for novel or rare compositional configurations at inference time. See additional details in App.~\ref{app:classifier_generalization}

\subsection{Inference-Time Composition Guidance}
\label{sec:inference_time_control}

Once the classifier is trained, we use it during denoising to steer generation toward the relation specified by the prompt. Given a prompt $p$ with target relation $y$, we initialize the diffusion process as usual and iteratively denoise the latent video variable. At a selected denoising step $t$, we extract the relevant subject-token cross-attention maps, compute the aggregated representation $\phi(z_t,p)$, and evaluate the classifier.

The classifier produces a score over all relations it has learned:
\begin{equation}
s_t = C(\phi(z_t,p)).
\end{equation}
We then define an inference-time cross-entropy loss against the target composition:
\begin{equation}
\mathcal{L}_{\mathrm{comp}}(z_t,p,y) = - \log s_t[y].
\end{equation}


This loss measures how strongly the current latent trajectory is recognized as expressing the desired composition. We then update the latent using the gradient step in Eq.~\ref{eq:latent_guidance_update}, backpropagating through the classifier and the extracted cross-attention maps.

Thus, during inference, the classifier receives the subject-token cross-attention maps, applies the learned aggregation module, produces a score for every composition class, and the latent is optimized by minimizing the cross-entropy with respect to the target composition. For the prompt ``a dog turning left,'' the classifier sees only the token \textit{dog}, predicts scores over all known compositional tokens, and the loss is computed against the target \textit{left}.

To preserve realism and avoid over-constraining the generation, we apply this guidance only during the first 8 steps, which are shown to be the steps where the coarse motion and spatial structure are formed~\cite{hertz2022prompt,liu2024subjectenhanced,feng2023trainingfree,shaulov2025flowmo}. 

\subsection{Inference-Time Multi-Composition Guidance}
\label{sec:multi_composition_guidance}

\ourmethod{} can also handle temporally structured compositional prompts, where different composition constraints should hold at different parts of the video, e.g., ``a dog turns left and then turns right.'' In this setting, the user provides a set of temporally localized composition constraints
\begin{equation}
\mathcal{Q}
=
\left\{
(y_i, \tau_i^{\mathrm{start}}, \tau_i^{\mathrm{end}})
\right\}_{i=1}^{M},
\end{equation}
where $y_i \in \mathcal{Y}$ is the target composition label and $\tau_i^{\mathrm{start}},\tau_i^{\mathrm{end}}$ define the time interval in which it should be expressed. Each interval is converted into a frame set $\Omega_i$. At a guided denoising step $t$, we compute the attention representation $\phi(z_t,p)$ and apply the classifier separately to each temporally masked segment,
\begin{equation}
s_{t,i}
=
C\!\left(\mathrm{Mask}_{\Omega_i}(\phi(z_t,p))\right).
\end{equation}
The multi-composition guidance loss is then
\begin{equation}
\mathcal{L}_{\mathrm{multi}}(z_t,p,\mathcal{Q})
=
\sum_{i=1}^{M}
\lambda_i
\left(-\log s_{t,i}[y_i]\right),
\end{equation}
and the latent is updated as
\begin{equation}
z_t'
=
z_t
-
\eta_t
\nabla_{z_t}
\mathcal{L}_{\mathrm{multi}}(z_t,p,\mathcal{Q}).
\end{equation}
Thus, each composition constraint supervises only the frames in which it is intended to occur, while the generator, classifier, and inference procedure remain unchanged. Additional details are provided in App.~\ref{app:multi_composition_guidance}.

\subsection{Inference-Time Multi-Composition Guidance in Autoregressive Video Generation}
\label{sec:ar_multi_composition_guidance}

\ourmethod{} also extends to autoregressive long-video generation, where a video is produced as a sequence of temporal batches. This setting naturally supports temporally evolving compositional prompts, such as ``a dog walks to the left of a bench, then moves behind it, and then stops in front of it.'' Rather than enforcing one global composition label over the full video, we assign a target composition to each generated batch.

Let the video be generated as $x=\{x^{(1)},\dots,x^{(M)}\}$, where $x^{(m)}$ is the $m$-th batch. We define batch-level composition constraints
\begin{equation}
\mathcal{Q}_{\mathrm{AR}}
=
\left\{(p_m,y_m)\right\}_{m=1}^{M},
\end{equation}
where $p_m$ is the prompt segment for batch $m$ and $y_m \in \mathcal{Y}$ is the target composition label. During guided denoising of batch $m$, we extract the subject-token cross-attention maps, compute $\phi(z_t^{(m)},p_m)$, and evaluate the classifier:
\begin{equation}
s_t^{(m)} = C(\phi(z_t^{(m)},p_m)).
\end{equation}
The batch-specific guidance loss is
\begin{equation}
\mathcal{L}_{\mathrm{AR}}^{(m)}
=
-\log s_t^{(m)}[y_m],
\end{equation}
and the current batch latent is updated by
\begin{equation}
z_t^{(m)\prime}
=
z_t^{(m)}
-
\eta_t
\nabla_{z_t^{(m)}}
\mathcal{L}_{\mathrm{AR}}^{(m)}.
\end{equation}
Thus, the compositional objective can change across autoregressive batches while temporal continuity is preserved through the generated context. Additional details are provided in App.~\ref{app:ar_multi_composition_guidance}.





\section{Experiments}
\label{Experiments}
We present a comprehensive evaluation of \ourmethod{} for composition-aware control in text-to-video generation. Our experiments examine whether a lightweight classifier trained on subject-token cross-attention maps can provide effective inference-time guidance for fine-grained compositional generation, without updating the weights of the underlying video generator. We evaluate \ourmethod{} in both short-video and long-horizon auto-regressive generation settings, testing its ability to improve composition accuracy while preserving visual quality, motion realism, and semantic alignment. We conduct quantitative evaluations, qualitative comparisons, and human preference studies across multiple text-to-video backbones and generation regimes.

\paragraph{\textbf{Implementation details.}\quad}


We evaluate \ourmethod{} on both short- and long-video generation. For short-video generation, we use Wan2.2~\cite{wan2025wan} and CogVideoX~\cite{yang2024cogvideox}; for long-horizon autoregressive generation, we use Rolling Forcing~\cite{liu2025rolling}. To obtain cross-attention maps for real videos, we first invert each video into the latent space of the frozen generator using the video inversion method of Yesiltepe et al.~\cite{yesiltepe2025dynamic}, and then replay the denoising trajectory to extract the corresponding attention maps. In all settings, \ourmethod{} operates purely at inference time: the pretrained text-to-video generator remains frozen, and guidance is applied by optimizing the latent video variable using the compositional-classifier loss. All experiments are conducted on a single NVIDIA H100 GPU. For Rolling Forcing, we generate 30-second videos at 16 FPS and a resolution of $832 \times 480$. For Wan2.2 and CogVideoX, we generate 5-second videos using the same frame rate and resolution. Additional details on classifier training, cross-attention extraction, dual inversion, and guidance scheduling are provided in App.~\ref{app:implementation_details}.

\begin{table}[h!]
\centering
\caption{\textbf{VBench summary for short- and long-video generation.} Comparison of base models with TTOM and \ourmethod{} for short videos, and Rolling Forcing with TTOM and \ourmethod{} for long videos.}
\label{tab:vbench_summary_short_long}
\setlength{\tabcolsep}{10pt}
\begin{tabular}{l c c c}
\toprule
\textbf{Model} & \textbf{Semantic} & \textbf{Quality} & \textbf{Final} \\
\midrule
\multicolumn{4}{c}{\textbf{Short-video generation}} \\
\midrule
CogVideoX-5B              & 73.2\% & 78.4\% & 75.8\% \\
+ TTOM                    & 75.1\% & 79.1\% & 77.1\% \\
+ \ourmethod{}              & \textbf{78.6\%} & \textbf{79.3\%} & \textbf{78.9\%} \\
\midrule
Wan2.2-5B                 & 75.8\% & 80.0\% & 77.9\% \\
+ TTOM                    & 77.6\% & 80.8\% & 79.2\% \\
+ \ourmethod{}              & \textbf{80.9\%} & \textbf{81.2\%} & \textbf{81.1\%} \\
\midrule
Wan2.2-14B                & 82.6\% & 84.6\% & 83.6\% \\
+ TTOM                    & 84.6\% & 85.4\% & 85.0\% \\
+ \ourmethod{}              & \textbf{89.2\%} & \textbf{85.6\%} & \textbf{86.3\%} \\
\midrule
\multicolumn{4}{c}{\textbf{Long-video generation}} \\
\midrule
Rolling Forcing           & 59.13\% & 74.37\% & 66.75\% \\
+ TTOM                    & 61.24\% & 74.92\% & 68.08\% \\
+ \ourmethod{}              & \textbf{63.82\%} & \textbf{75.41\%} & \textbf{69.62\%} \\
\bottomrule
\end{tabular}
\end{table}

\begin{table*}[t]
\centering
\caption{\textbf{Composition-aware evaluation across all dimensions.} Comparison of base models with TTOM and \ourmethod{} on T2V-CompBench~\cite{sun2025t2v}.}
\label{tab:relation_eval_all_models}
\setlength{\tabcolsep}{5pt}
\begin{tabular*}{\textwidth}{@{\extracolsep{\fill}} l c c c c c c c c}
\toprule
\textbf{Model} &
\textbf{Motion} &
\textbf{Num} &
\textbf{Spatial} &
\textbf{Con-attr} &
\textbf{Dyn-attr} &
\textbf{Action} &
\textbf{Interact} &
\textbf{Avg} \\
\midrule
CogVideoX-5B       & 0.266 & 0.371 & 0.517 & 0.616 & 0.022 & 0.533 & 0.607 & 0.419 \\
+TTOM              & 0.332 & 0.392 & 0.566 & 0.742 & 0.029 & \textbf{0.671} & \textbf{0.638} & 0.482 \\
+\textbf{\ourmethod{}} & \textbf{0.412} & \textbf{0.398} & \textbf{0.624} & \textbf{0.786} & \textbf{0.037} & 0.663 & 0.622 & \textbf{0.506} \\
\midrule
Wan2.2-5B          & 0.248 & 0.462 & 0.533 & 0.781 & 0.051 & 0.704 & 0.671 & 0.486 \\
+TTOM              & 0.287 & 0.474 & 0.561 & 0.804 & 0.063 & 0.731 & \textbf{0.689} & 0.507 \\
+\textbf{\ourmethod{}} & \textbf{0.338} & \textbf{0.478} & \textbf{0.603} & \textbf{0.816} & \textbf{0.073} & \textbf{0.735} & 0.682 & \textbf{0.532} \\
\midrule
Wan2.2-14B         & 0.301 & 0.521 & 0.592 & 0.857 & 0.091 & 0.792 & 0.736 & 0.556 \\
+TTOM              & 0.348 & 0.533 & 0.618 & 0.872 & 0.102 & \textbf{0.806} & \textbf{0.751} & 0.579 \\
+\textbf{\ourmethod{}} & \textbf{0.417} & \textbf{0.535} & \textbf{0.667} & \textbf{0.888} & \textbf{0.117} & 0.795 & 0.742 & \textbf{0.594} \\
\bottomrule
\end{tabular*}
\end{table*}

\subsection{Quantitative Results}
\label{quantitative_results}

We evaluate \ourmethod{} using both composition-specific and general-purpose video generation benchmarks. For compositional faithfulness, we report performance on T2V-CompBench~\cite{sun2025t2v}, which measures motion, numeracy, constitutionality, content attributes, dynamic attributes, actions, and interactions. To evaluate whether composition guidance preserves overall generation quality, we further report VBench~\cite{huang2024vbench} for short videos and VBench-Long~\cite{huang2024vbenchpp} for long-horizon autoregressive generation. We compare against the corresponding base generators and TTOM~\cite{qu2025ttom}, a strong layout-guided test-time optimization baseline, using identical prompts and generation settings.

\paragraph{\textbf{Automatic metrics.}\quad}
Tab.~\ref{tab:vbench_summary_short_long} shows that \ourmethod{} consistently improves over both the base generators and TTOM. For short-video generation, \ourmethod{} achieves the best \textbf{Final Score} on every backbone: 78.9\% on CogVideoX-5B, 81.1\% on Wan2.2-5B, and 86.3\% on Wan2.2-14B. In all cases, it outperforms the base model and the TTOM baseline. The gains are mainly driven by stronger semantic alignment, while the \textbf{Quality Score} remains stable or slightly improves, indicating that our guidance improves prompt faithfulness without degrading the visual prior of the underlying generator.

Tab.~\ref{tab:relation_eval_all_models} further evaluates composition-aware generation on T2V-CompBench. \ourmethod{} achieves the best average score on every backbone, improving CogVideoX-5B from 0.419 to 0.506, Wan2.2-5B from 0.486 to 0.532, and Wan2.2-14B from 0.556 to 0.594. Importantly, \ourmethod{} also consistently outperforms TTOM, despite TTOM relying on layout-based guidance, while our method requires no user-provided layouts, boxes, or spatial controls. The gains are concentrated in composition-sensitive dimensions such as \textbf{Motion}, \textbf{Spatial}, \textbf{Con-attr}, and \textbf{Dyn-attr}, supporting the effectiveness of subject-token cross-attention as a guidance signal for fine-grained compositional control.

The same trend holds in long-horizon autoregressive generation. On Rolling Forcing, \ourmethod{} improves the \textbf{Final Score} from 66.75\% to 69.62\%, outperforming TTOM at 68.08\%. It also improves both \textbf{Semantic Score} and \textbf{Quality Score}, suggesting that composition-aware guidance remains effective beyond short clips, where maintaining compositional consistency over time is more challenging. Additional per-dimension results, including \textbf{Spatial Relationship} and \textbf{Multiple Objects}, are provided in App.~\ref{app:VBench_Metrics_Breakdown} and App.~\ref{app:VBench_long_Metrics_Breakdown}.

\subsection{Qualitative Results}
\label{sec:qualitative_res}
\begin{figure*}[t]
  \centering
  \includegraphics[width=0.9\textwidth]{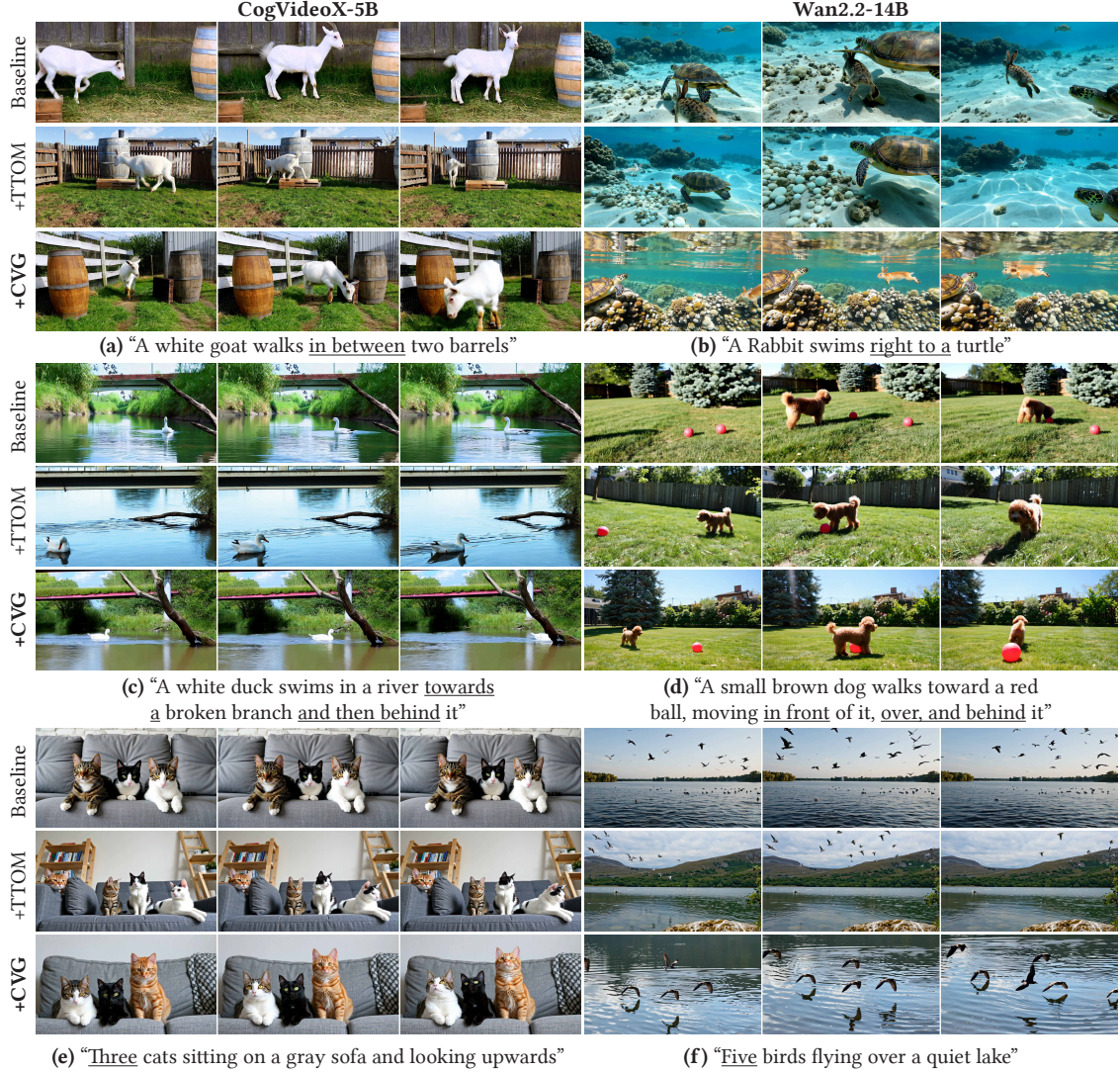}
  \caption{\textbf{Qualitative results.} Text-to-video results before and after applying CVG on Wan2.2-14B~\cite{wan2025wan} and CogVideoX-5B~\cite{hong2022cogvideo}. CVG strictly adheres to compositional instructions.}
  \label{fig:qualitative_cvg}
\end{figure*}

We complement the quantitative evaluation with a qualitative analysis on prompts drawn from T2V-CompBench~\cite{sun2025t2v}, covering the full range of compositional dimensions reported in Sec.~\ref{quantitative_results}, directional motion, spatial relations, numeracy, action and instruction binding, and multi-relation temporal composition. Comparisons are made against the corresponding baselines, CogVideoX-5B~\cite{yang2024cogvideox} and Wan2.2-14B~\cite{wang2025wan}, as well as the inference-time optimization baseline TTOM~\cite{qu2025ttom}, using identical prompts and generation settings. Representative examples are shown in Fig.~\ref{fig:teaser} and Fig.~\ref{fig:qualitative_cvg}, with multi-relation cases highlighted in Fig.~\ref{fig:qualitative_cvg}(c,~d).

Across these examples, CVG produces videos that faithfully follow the compositional structure of the prompt, whereas the baselines and TTOM frequently violate one or more of the specified constraints. We observe three recurring failure modes. First, \textbf{incorrect directional and topological motion}: in Fig.~\ref{fig:teaser}(a), Wan2.2-14B places the rabbit beside rather than \emph{on top of} the turtle, and in Fig.~\ref{fig:teaser}(b), CogVideoX-5B depicts the crab on the open sand rather than crawling \emph{from beneath} the driftwood. Second, \textbf{mislocalized spatial relations}: in Fig.~\ref{fig:qualitative_cvg}(a), the baselines do not place the goat \emph{in between} the two barrels, and in Fig.~\ref{fig:qualitative_cvg}(b), both Wan2.2-14B and TTOM fail to position the rabbit to the \emph{right of} the turtle. Third, \textbf{miscounted entities and unstable instruction binding}: in Fig.~\ref{fig:qualitative_cvg}(e), the baselines either generate the wrong number of cats or fail to enforce the \emph{looking upwards} instruction, and in Fig.~\ref{fig:qualitative_cvg}(f), Wan2.2-14B yields an inconsistent number of birds across frames.

Compositional prompts that combine \textbf{multi-stage motion} further amplify the gap. In Fig.~\ref{fig:qualitative_cvg}(c), CogVideoX-5B never moves the duck towards the branch, and TTOM partially establishes the \emph{towards} phase but fails to follow it with the \emph{behind} phase; CVG instead produces a coherent trajectory in which the duck first swims towards the branch and then disappears behind it. A similar pattern appears in Fig.~\ref{fig:qualitative_cvg}(d), where Wan2.2-14B and TTOM generate a dog that drifts ambiguously around the ball, while CVG faithfully follows all three relations -- \emph{in front of}, \emph{over}, and \emph{behind} -- as distinct, temporally ordered phases.

These examples illustrate that classifier-guided steering over subject-token cross-attention enforces fine-grained compositional structure -- directionality, counting, instruction grounding, and temporal ordering of relations -- while preserving the appearance and motion priors of the underlying generator.

Additional qualitative comparisons across further directional, spatial, numerical, and action-binding prompts are provided in the App~\ref{sec:additional_qualitative}.

\begin{figure*}[t]
  \centering
  \includegraphics[width=0.9\textwidth]{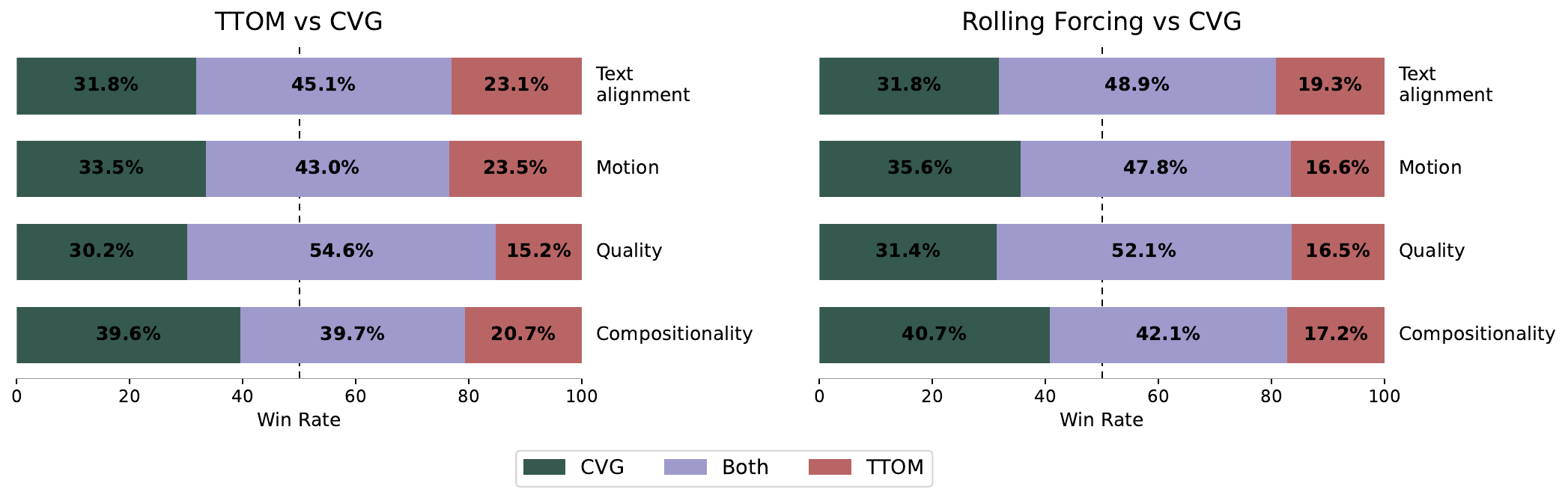}
  \caption{\textbf{User study results.} Human preference comparison between \ourmethod{} and TTOM. Participants compare generated videos along text alignment, motion, quality, and compositionality, and select whether \ourmethod{}, TTOM, or both are better.}
  \label{fig:user_study_ttom_vs_ours}
\end{figure*}

\subsection{User Study}
\label{sec:userstudy}

We further conduct a human preference study to evaluate whether the quantitative gains of \ourmethod{} translate into perceptual improvements in generated videos. We randomly select 100 prompts from T2V-CompBench and generate videos using the same prompts and generation settings for both TTOM and \ourmethod{}. For each prompt, participants are shown two generated videos, one produced with TTOM and one produced with \ourmethod{}, in randomized order. They are asked to indicate whether \ourmethod{} is better, TTOM is better, or both are comparable along four criteria: \textbf{Text Alignment}, \textbf{Motion}, \textbf{Quality}, and \textbf{Compositionality}. The last criterion measures whether the generated video correctly realizes the fine-grained compositional constraints in the prompt, such as directional motion, spatial relations, multi-object interactions, and attribute binding.

Fig.~\ref{fig:user_study_ttom_vs_ours} summarizes the results. On Wan2.2-14B, \ourmethod{} is preferred over TTOM in all evaluated criteria. In particular, \ourmethod{} is preferred for \textbf{Compositionality} in 39.6\% of comparisons, while TTOM is preferred in only 20.7\%. A similar trend is observed for \textbf{Text Alignment} and \textbf{Motion}, where \ourmethod{} is preferred in 31.8\% and 33.5\% of comparisons, respectively, compared to 23.1\% and 23.5\% for TTOM. For \textbf{Quality}, most comparisons are judged comparable, with 54.6\% marked as both, indicating that the proposed guidance improves compositional faithfulness without noticeably degrading visual quality.

The same pattern holds in the long-video setting with Rolling Forcing. \ourmethod{} is preferred for \textbf{Compositionality} in 40.7\% of comparisons, compared to 17.2\% for TTOM. It is also preferred for \textbf{Motion} in 35.6\% of comparisons, compared to 16.6\% for TTOM, and for \textbf{Text Alignment} in 31.8\% of comparisons, compared to 19.3\% for TTOM. Again, \textbf{Quality} receives the largest fraction of ``both'' votes, with 52.1\% of comparisons judged comparable, suggesting that \ourmethod{} preserves the appearance quality of the underlying generator while improving the realization of compositional constraints.

Overall, the user study supports the quantitative findings in Sec.~\ref{Experiments}. Human raters consistently prefer \ourmethod{} over TTOM for compositional correctness, text alignment, and motion, while visual quality is largely preserved. These results indicate that using a relation-aware classifier as an inference-time guidance objective provides a perceptually meaningful improvement over generic test-time optimization.

\section{Limitations \& Future Work}



\ourmethod{} operates purely at inference time and leaves the underlying video generator unchanged; therefore, it improves compositional faithfulness by steering the denoising trajectory rather than learning new generative capabilities. Although the VLM backbone enables the classifier to generalize to unseen composition labels, its effectiveness still depends on the quality of the model’s internal grounding signals and on how well the requested composition is reflected in cross-attention. The current formulation is strongest for subject-centric compositions such as motion direction, spatial layout, and attribute binding, and may be less reliable for ambiguous prompts, subtle compositions, or scenes with multiple interacting subjects. Looking forward, these limitations point to a broader opportunity: turning internal representations of pretrained video generators into general-purpose control interfaces. Future work could extend this direction toward structured multi-entity composition graphs, stronger language-conditioned objectives, and automatic temporal planning for complex prompts such as ``first turn left, then move away.’’



\section{Conclusion}

In this work, we address compositional failures in text-to-video generation by using the model’s internal grounding signals as an inference-time control interface, rather than relying on external controls, layouts, or video-scale fine-tuning. \ourmethod{} leverages entity-token cross-attention maps to train a lightweight compositional classifier, using video inversion and denoising replay to obtain these maps from real videos. To prevent composition leakage, we apply dual inversion with correct and incorrect composition prompts while supervising both with the same ground-truth label. At inference time, the classifier provides a differentiable guidance signal that steers early denoising steps toward the desired composition, improving compositional accuracy while preserving visual quality. Across short- and long-video generation, multiple T2V models, automatic benchmarks, and human studies, our results show that compositional faithfulness can be improved without modifying model weights. More broadly, our findings suggest that internal representations of video diffusion models can serve as actionable control mechanisms for aligning generated videos with complex compositional prompts.

\bibliographystyle{splncs04}
\bibliography{references}



\clearpage
\appendix

\section{VBench Metrics Breakdown}
\label{app:VBench_Metrics_Breakdown}

We provide the full per-dimension VBench results for short-video generation in
Tabs.~\ref{tab:vbench_cogvideox_5b_full},
\ref{tab:vbench_wan22_5b_full}, and
\ref{tab:vbench_wan22_14b_full}. We compare each base generator with TTOM and
\ourmethod{} under identical prompts and generation settings. Across all three
backbones, \ourmethod{} improves the aggregate VBench scores while preserving
the visual quality of the underlying generator. The gains are especially
pronounced in semantic and composition-sensitive dimensions, indicating that
cross-attention guidance improves prompt faithfulness rather than merely
changing low-level appearance.

For \textbf{CogVideoX-5B}, \ourmethod{} improves the \textbf{Semantic Score}
from 73.2\% to 78.6\% and the \textbf{Final Score} from 75.8\% to 78.8\%,
outperforming TTOM, which obtains a Final Score of 77.1\%. The largest gains
appear in composition-related dimensions: \textbf{Spatial Relationship} improves
by +10.9\%, \textbf{Dynamic Degree} by +6.7\%, \textbf{Human Action} by +4.3\%,
and \textbf{Multiple Objects} by +3.6\%. These improvements suggest that
\ourmethod{} strengthens spatial grounding and motion-related composition while
maintaining comparable visual quality.

A similar trend is observed for \textbf{Wan2.2-5B}. \ourmethod{} increases the
\textbf{Semantic Score} from 75.8\% to 80.9\% and the \textbf{Final Score} from
77.9\% to 80.5\%, again outperforming TTOM. The strongest improvements are
obtained in \textbf{Spatial Relationship} (+10.1\%), \textbf{Dynamic Degree}
(+5.9\%), \textbf{Human Action} (+4.4\%), and \textbf{Multiple Objects}
(+3.7\%). Notably, the \textbf{Quality Score} also increases from 80.0\% to
81.2\%, indicating that the proposed guidance improves compositional alignment
without sacrificing visual fidelity.

For the stronger \textbf{Wan2.2-14B} backbone, \ourmethod{} improves the
\textbf{Semantic Score} from 82.6\% to 89.2\% and the \textbf{Final Score} from
83.6\% to 86.3\%, outperforming TTOM at 85.0\%. The largest gains again occur in
composition-sensitive dimensions, including \textbf{Spatial Relationship}
(+12.2\%), \textbf{Dynamic Degree} (+6.8\%), \textbf{Human Action} (+4.9\%), and
\textbf{Multiple Objects} (+4.8\%). These results show that \ourmethod{} remains
effective even when applied to a stronger base generator, suggesting that
subject-token cross-attention provides complementary compositional control beyond
the generator's native capabilities.

Overall, the full VBench breakdown shows that \ourmethod{} consistently improves
semantic alignment and compositional faithfulness across different T2V
backbones. While TTOM occasionally improves appearance-oriented dimensions such
as background consistency or aesthetic quality, \ourmethod{} achieves stronger
and more consistent gains in composition-related metrics and obtains the best
aggregate performance across all evaluated models. This supports our central
claim that internal cross-attention maps can serve as an effective inference-time
control signal for fine-grained compositional video generation.

\begin{table*}[h!]
\centering
\caption{\textbf{VBench evaluation across all dimensions for CogVideoX-5B.} Comparison of the base model with TTOM and \ourmethod{}.}
\label{tab:vbench_cogvideox_5b_full}
\setlength{\tabcolsep}{5pt}
\begin{tabular*}{\textwidth}{@{\extracolsep{\fill}} l c c c}
\toprule
\textbf{Dimension} &
\textbf{CogVideoX-5B} &
\textbf{+ TTOM} &
\textbf{+ \ourmethod{}} \\
\midrule
Subject Consistency        & 82.1\% & 83.0\% & \textbf{84.2\%} \\
Background Consistency     & 80.3\% & \textbf{80.9\%} & 80.7\% \\
Temporal Flickering        & 98.4\% & 98.6\% & \textbf{98.8\%} \\
Motion Smoothness          & 95.1\% & 95.9\% & \textbf{96.8\%} \\
Dynamic Degree             & 50.4\% & 52.3\% & \textbf{57.1\%} \\
Aesthetic Quality          & 60.8\% & \textbf{61.4\%} & 61.1\% \\
Imaging Quality            & 63.2\% & 63.8\% & \textbf{64.0\%} \\
Object Class               & 78.4\% & 79.1\% & \textbf{80.6\%} \\
Multiple Objects           & 56.8\% & 58.1\% & \textbf{60.4\%} \\
Human Action               & 74.3\% & 75.8\% & \textbf{78.6\%} \\
Color                      & 80.1\% & 80.6\% & \textbf{81.2\%} \\
Spatial Relationship       & 61.4\% & 65.8\% & \textbf{72.3\%} \\
Scene                      & 42.6\% & \textbf{43.1\%} & 42.6\% \\
Appearance Style           & 24.7\% & \textbf{25.4\%} & 25.1\% \\
Temporal Style             & 19.8\% & 20.2\% & \textbf{20.9\%} \\
Overall Consistency        & 24.6\% & 25.1\% & \textbf{25.9\%} \\
\midrule
Semantic Score             & 73.2\% & 75.1\% & \textbf{78.6\%} \\
Quality Score              & 78.4\% & 79.1\% & \textbf{79.3\%} \\
Final Score                & 75.8\% & 77.1\% & \textbf{78.8\% \textcolor{green!70!black}{(+3.0\%)}} \\
\bottomrule
\end{tabular*}
\end{table*}

\section{VBench Long Metrics Breakdown}
\label{app:VBench_long_Metrics_Breakdown}

We provide the full per-dimension VBench-Long results for long-horizon autoregressive generation in Tab.~\ref{tab:vbench_long_rolling_forcing_worldcontrol}. We compare Rolling Forcing with TTOM and \ourmethod{} under identical prompts and generation settings. The results show that \ourmethod{} consistently improves the aggregate scores while maintaining the visual quality of the underlying autoregressive generator. For \textbf{Rolling Forcing}, \ourmethod{} improves the \textbf{Semantic Score} from 59.13\% to 63.82\% and the \textbf{Final Score} from 66.75\% to 69.62\%, outperforming TTOM, which obtains a Final Score of 68.08\%. The strongest gains appear in composition- and motion-sensitive dimensions, including \textbf{Spatial Relationship} (+5.95\%), \textbf{Dynamic Degree} (+5.69\%), \textbf{Human Action} (+4.62\%), and \textbf{Motion Smoothness} (+1.68\%). These improvements indicate that \ourmethod{} enhances long-horizon compositional alignment and motion structure, rather than only improving local visual appearance.Compared to TTOM, \ourmethod{} also provides stronger gains in the dimensions most related to compositional control. In particular, \ourmethod{} improves over TTOM in \textbf{Spatial Relationship} (81.37\% vs. 78.46\%), \textbf{Dynamic Degree} (56.93\% vs. 53.41\%), \textbf{Human Action} (76.84\% vs. 74.05\%), and \textbf{Motion Smoothness} (92.18\% vs. 91.36\%). While TTOM slightly improves a small number of appearance-oriented metrics, such as \textbf{Imaging Quality} and \textbf{Scene}, \ourmethod{} achieves the best aggregate performance, with higher \textbf{Semantic Score}, \textbf{Quality Score}, and \textbf{Final Score}. Overall, the VBench-Long breakdown shows that the proposed guidance remains effective beyond short clips. By applying composition-aware control during autoregressive generation, \ourmethod{} improves semantic alignment, spatial grounding, and motion consistency over long temporal rollouts while preserving video quality.

\section{Training Composition Labels}
\label{app:Training_Composition_Labels}
The compositional classifier used by \ourmethod{} is trained on a limited set of composition labels, but is designed to generalize beyond this closed training vocabulary. Specifically, during training we supervise the classifier using the following composition classes:
\[
\begin{aligned}
\mathcal{Y}_{\mathrm{train}}
=
\{&
\textit{above},
\textit{away},
\textit{behind},
\textit{beneath},\\
&
\textit{in\_front\_of},
\textit{inside},
\textit{next\_to},
\textit{toward}
\}.
\end{aligned}
\]
These labels cover a compact set of primitive subject-centric relations, including vertical relations, depth ordering, containment, proximity, and directional motion. Although this vocabulary is limited, it provides supervision over fundamental spatial and motion patterns that appear repeatedly in more complex compositional prompts.

\section{Additional implementation Details For Compositional Classifier}
\label{Additional_Implementation_Details_For_Compositional_classifier}

This appendix provides additional details on the compositional classifier used by \ourmethod{}. The classifier is trained to predict composition labels from the internal cross-attention maps of a frozen text-to-video generator. Unlike the main paper, which summarizes the classifier at a high level, we describe here the cross-attention extraction procedure, the aggregation architecture, the real and synthetic training data, the leakage-handling strategy, and the inference-time guidance implementation.

\subsection{Details For Cross-Attention as a Composition Representation}
\label{Additional_Implementation_Details_cross_attenstion}

A central design choice in \ourmethod{} is to use cross-attention maps from the pretrained text-to-video diffusion model as the representation on which control is learned. Cross-attention captures how the model grounds prompt tokens in the evolving spatiotemporal latent, making it a natural signal for inferring subject-centric relations.

To obtain these cross-attention maps from real videos, we first apply \emph{video inversion}. Given a real video and its associated prompt, we invert the video into the latent space of the frozen text-to-video model, and then replay the denoising trajectory to extract the corresponding cross-attention maps. This allows us to align real video supervision with the model's internal representations, and to train the classifier directly on the same type of attention signals that will be used later during inference-time control.

Let $A_{t,\ell,h}^{(k)} \in \mathbb{R}^{F \times H \times W}$ denote the cross-attention map corresponding to token $k$ at denoising step $t$, layer $\ell$, and attention head $h$, where $F$ is the number of frames and $H \times W$ is the latent spatial resolution. Given a prompt $p$, we extract only the cross-attention maps corresponding to the \emph{composition-relevant entity tokens}, rather than the relation token itself. For example, for the prompt ``a dog turning left,'' we use the cross-attention maps of the token \textit{dog}. For a two-object prompt such as ``a dog moves behind a ball,'' we extract the maps of both \textit{dog} and \textit{ball}. For a three-object prompt such as ``a dog stands between a ball and a bench,'' we extract the maps of \textit{dog}, \textit{ball}, and \textit{bench}. Thus, the classifier observes the grounded entities involved in the composition, while the target relation or composition label is withheld from its input.

Let $\mathcal{K}(p)$ denote the set of composition-relevant entity tokens in prompt $p$. For each selected denoising step $t \in \mathcal{T}$, we collect all corresponding cross-attention maps across the chosen layers $\mathcal{L}$ and heads $\mathcal{H}$:
\begin{equation}
\mathcal{S}_t(p)=\left\{A_{t,\ell,h}^{(k)} \;:\; \ell \in \mathcal{L},\; h \in \mathcal{H},\; k \in \mathcal{K}(p)\right\}.
\end{equation}

Notably, the classifier does not receive the composition token itself. Instead, it predicts the target composition solely from the temporal evolution of the composition-relevant entity-token cross-attention maps.

We retain the spatiotemporal structure of the extracted attention maps and aggregate them using a learned module. For a fixed denoising step $t$, each attention map in $\mathcal{S}_t(p)$ is first flattened into a token:
\begin{equation}
v_{t,\ell,h}^{(k)} = \mathrm{vec}\!\left(A_{t,\ell,h}^{(k)}\right) \in \mathbb{R}^{FHW}.
\end{equation}
Each token is then projected into a shared embedding space of dimension $d$:
\begin{equation}
e_{t,\ell,h}^{(k)} = W_{\mathrm{in}} v_{t,\ell,h}^{(k)} + b_{\mathrm{in}} \in \mathbb{R}^{d}.
\end{equation}
We enrich each token with a layer embedding $e^{\mathrm{layer}}_{\ell}$, a head embedding $e^{\mathrm{head}}_{h}$, and a denoising-step embedding $e^{\mathrm{time}}_{t}$:
\begin{equation}
\hat{e}_{t,\ell,h}^{(k)} = e_{t,\ell,h}^{(k)} + e^{\mathrm{layer}}_{\ell} + e^{\mathrm{head}}_{h} + e^{\mathrm{time}}_{t}.
\end{equation}


The resulting sequence is processed by a trainable transformer encoder:
\begin{equation}
U_t = \mathrm{TransformerEnc}_{\theta_{\mathrm{agg}}}\!\left(\{\hat{e}_{t,\ell,h}^{(k)}\}\right),
\end{equation}
where $U_t \in \mathbb{R}^{N_t \times d}$ and $N_t = |\mathcal{L}|\,|\mathcal{H}|\,|\mathcal{K}(p)|$.

To aggregate the sequence into a single spatiotemporal descriptor, we introduce a learned query vector $q \in \mathbb{R}^{d}$ and apply cross-attention from $q$ to the encoded sequence:
\begin{equation}
\hat{u}_t = \mathrm{CrossAttn}(q, U_t) \in \mathbb{R}^{d}.
\end{equation}
Finally, we project the aggregated embedding back to the spatiotemporal attention domain:
\begin{equation}
\hat{A}_t = \mathrm{reshape}\!\left(W_{\mathrm{out}} \hat{u}_t + b_{\mathrm{out}}\right) \in \mathbb{R}^{F \times H \times W}.
\end{equation}

Thus, for each selected denoising step, the aggregation module outputs a single spatiotemporal attention volume $\hat{A}_t$ summarizing how the relevant entities are grounded across time and space. We then concatenate the aggregated volumes across the selected denoising steps:
\begin{equation}
\phi(z_t,p) = \mathrm{Concat}\!\left(\hat{A}_{t_1}, \hat{A}_{t_2}, \dots, \hat{A}_{t_{|\mathcal{T}|}}\right)
\in \mathbb{R}^{|\mathcal{T}| \times F \times H \times W}.
\end{equation}

This representation is fed to a frozen VLM backbone $B$, followed by a trainable classification head:
\begin{equation}
h = B(\phi(z_t,p)),
\end{equation}
\begin{equation}
C(\phi(z_t,p)) = \mathrm{softmax}(W h + b) \in \Delta^{|\mathcal{Y}|},
\end{equation}
where $W$ and $b$ are the parameters of the classification head.

\subsection{Details For Compositional classifier Training}
\label{Compositional_Classifier_Additional_Implementation_Details}

\paragraph{\textbf{Training Data: Real and Synthetic Images.}}
The classifier is trained on a combination of real and synthetic data. For real data, we use VidOR~\cite{shang2019annotating} and ImageNet-VidVRD~\cite{shang2017video}, which provide supervision for temporal and motion-aware relations. For synthetic data, we use the dataset introduced in \emph{VChain} ~\cite{huang2025vchain}. For each training example, we construct a prompt, extract the subject-token cross-attention maps from the frozen generator, and pair the resulting representation with a ground-truth composition label $y$. The real video datasets expose the classifier to natural motion patterns and object interactions, while the synthetic data increases coverage and diversity of compositions configurations.

To reduce \emph{composition leakage}~\cite{yiflach2025data}, we adopt a video analogue of dual inversion. \emph{Composition leakage} is defined as a shortcut in which the classifier predicts the target composition from textual traces of the relation in the attention maps, rather than from the visual spatiotemporal arrangement of the grounded entities. For each training video, we invert the same video twice: once with a prompt containing the correct composition, and once with a prompt containing an incorrect composition sampled from the label vocabulary. Let $p^{+}$ denote the prompt with the correct composition and $p^{-}$ the prompt with the incorrect composition. We extract subject-token cross-attention maps from both inversions, yielding two representations $\phi(z_t,p^{+})$ and $\phi(z_t,p^{-})$, but supervise both using the same ground-truth composition label $y$:
\begin{equation}
\mathcal{L}_{\mathrm{cls}}
=
-\log C(\phi(z_t,p^{+}))_y
-\log C(\phi(z_t,p^{-}))_y .
\end{equation}
This prevents the classifier from exploiting textual traces of the relation word in the attention maps, and instead forces it to rely on the actual spatiotemporal grounding of the subject. During training, the text-to-video generator and the VLM backbone remain frozen, and only the aggregation module parameters $\theta_{\mathrm{agg}}$ and the classification head parameters $(W,b)$ are optimized.

\paragraph{\textbf{Cross-Attention Extraction.}}
For each selected denoising step $t \in \mathcal{T}$, we collect the subject-token cross-attention maps across the chosen layers $\mathcal{L}$ and heads $\mathcal{H}$. Each attention map is represented as
\[
A_{t,\ell,h}^{(k)} \in \mathbb{R}^{F \times H \times W},
\]
where $F$ is the number of frames and $H \times W$ is the latent spatial resolution. We use only the cross-attention maps corresponding to the subject token, e.g., for the prompt ``a dog turning left,'' we extract only the maps associated with \textit{dog}.

\paragraph{\textbf{Classifier Architecture.}}
The relation classifier is composed of a learned aggregation module, a frozen Qwen2.5-VL-7B backbone, and a trainable linear classification head. The aggregation module first flattens each subject-token attention map and projects it to a shared embedding space. We add learnable layer, head, and denoising-step embeddings, and process the resulting sequence with a lightweight transformer encoder. A learned query token pools the encoded sequence into a single feature, which is projected back into a spatiotemporal tensor and concatenated across denoising steps. The resulting representation is fed into the frozen Qwen2.5-VL-7B backbone, followed by a trainable classification head that outputs logits over the relation label space.

\paragraph{\textbf{Training Setup.}}
During classifier training, the text-to-video generator and the Qwen2.5-VL-7B backbone remain frozen. Only the aggregation module and the classification head are optimized. We train using AdamW with learning rate \texttt{[LR]}, weight decay \texttt{[WD]}, batch size \texttt{[BATCH]}, and for \texttt{[EPOCHS]} epochs. We use mixed-precision training and a cosine learning-rate schedule with \texttt{[WARMUP]} warmup steps. Unless stated otherwise, gradients are propagated only through the aggregation module and classification head.

\paragraph{\textbf{Inference-Time Guidance.}}
At inference time, given a prompt with target relation $y$, we extract the subject-token cross-attention maps at the selected denoising steps, apply the learned aggregation module, and evaluate the frozen Qwen2.5-VL-7B backbone and trainable classification head to obtain relation logits. We then compute the cross-entropy loss against the target relation and backpropagate it to the current latent:
\[
\mathcal{L}_{\mathrm{rel}}(z_t,p,y) = -\log C(\phi(z_t,p))_y.
\]
Guidance is applied only at the first 8 denoising steps, where coarse relation and motion structure are formed~\cite{shaulov2025flowmo}. The final denoising steps are left unchanged so that the pretrained generator can refine appearance and fine details naturally.

\paragraph{\textbf{Implementation.}}
We implement \ourmethod{} in PyTorch. The text-to-video generator is used in frozen mode for both inversion and inference-time guidance. Qwen2.5-VL-7B is loaded as a frozen backbone, and only the aggregation transformer and classification head are trained. All experiments are run with mixed precision.

\begin{table*}[h!]
\centering
\caption{\textbf{VBench evaluation across all dimensions for Wan2.2-5B.} Comparison of the base model with TTOM and \ourmethod{}.}
\label{tab:vbench_wan22_5b_full}
\setlength{\tabcolsep}{5pt}
\begin{tabular*}{\textwidth}{@{\extracolsep{\fill}} l c c c}
\toprule
\textbf{Dimension} &
\textbf{Wan2.2-5B} &
\textbf{+ TTOM} &
\textbf{+ \ourmethod{}} \\
\midrule
Subject Consistency        & 83.9\% & 84.8\% & \textbf{85.8\%} \\
Background Consistency     & 82.4\% & \textbf{83.0\%} & 82.8\% \\
Temporal Flickering        & 98.7\% & 98.8\% & \textbf{99.0\%} \\
Motion Smoothness          & 95.7\% & 96.4\% & \textbf{97.2\%} \\
Dynamic Degree             & 53.1\% & 55.0\% & \textbf{59.0\%} \\
Aesthetic Quality          & 62.1\% & \textbf{62.8\%} & 62.5\% \\
Imaging Quality            & 64.5\% & \textbf{65.1\%} & 64.9\% \\
Object Class               & 80.2\% & 81.0\% & \textbf{82.5\%} \\
Multiple Objects           & 58.4\% & 59.7\% & \textbf{62.1\%} \\
Human Action               & 76.2\% & 77.9\% & \textbf{80.6\%} \\
Color                      & 81.5\% & 82.0\% & \textbf{82.8\%} \\
Spatial Relationship       & 64.7\% & 68.4\% & \textbf{74.8\%} \\
Scene                      & 44.1\% & 44.8\% & \textbf{45.7\%} \\
Appearance Style           & 25.4\% & \textbf{26.1\%} & 25.8\% \\
Temporal Style             & 20.5\% & 20.9\% & \textbf{21.6\%} \\
Overall Consistency        & 25.2\% & 25.8\% & \textbf{26.6\%} \\
\midrule
Semantic Score             & 75.8\% & 77.6\% & \textbf{80.9\%} \\
Quality Score              & 80.0\% & 80.8\% & \textbf{81.2\%} \\
Final Score                & 77.9\% & 79.2\% & \textbf{80.5\% \textcolor{green!70!black}{{(+2.6\%)}}} \\
\bottomrule
\end{tabular*}
\end{table*}

\section{Generalization Beyond the Training Composition Vocabulary}
\label{app:classifier_generalization}


A key reason for this generalization is that the classifier is not trained directly on raw pixels or on the text prompt alone. Instead, it receives subject-token cross-attention maps extracted from the frozen text-to-video generator. These maps encode how the model grounds the subject over space and time, and therefore expose geometric and temporal cues that are shared across many different compositions. For example, the distinction between \textit{toward} and \textit{away} depends on the temporal displacement of the subject grounding, while relations such as \textit{behind}, \textit{in\_front\_of}, and \textit{beneath} depend on consistent spatial organization across frames. Learning these primitive relations encourages the classifier to capture reusable compositional features rather than memorizing individual labels.

In addition, the classifier is built on top of a frozen VLM backbone, which provides a strong semantic and spatial prior from large-scale vision-language pretraining. As a result, the learned classifier head is not the only source of compositional knowledge. The VLM backbone already encodes broad visual concepts such as object count, relative position, containment, occlusion, and interaction. The supervised composition labels therefore act as anchors that align the cross-attention representation with this broader prior. This allows the classifier to extrapolate to compositions that were not explicitly observed during training.

This property is particularly useful for prompts containing relations outside the supervised label set. For instance, although the classifier is not explicitly trained with an \textit{in\_between} label, this relation can be partially inferred from the learned primitives \textit{next\_to}, \textit{in\_front\_of}, and \textit{behind}, together with the VLM's prior over multi-object spatial configurations. Similarly, numerical prompts such as ``three dogs beside a bench'' or ``two birds above a tree'' rely on object multiplicity and grounding consistency, which are not direct training labels but are represented in the VLM backbone and reflected in the cross-attention structure. Thus, even when the requested composition is not part of $\mathcal{Y}_{\mathrm{train}}$, the classifier can still provide a meaningful guidance signal by mapping the prompt to related spatial or semantic patterns.

At inference time, this enables \ourmethod{} to support a broader class of compositional instructions than the finite set used for classifier supervision. When the target composition corresponds exactly to one of the trained labels, we use the corresponding class directly. For novel compositions, we map the requested instruction to the closest supported primitive or combination of primitives. For example, \textit{between} can be treated as a multi-object spatial constraint involving proximity to two reference objects, while numerical relations can be guided through the subject grounding and semantic consistency captured by the VLM representation. In practice, this provides useful gradients for compositions such as \textit{between}, multi-object spatial layouts, and counting-related prompts, despite the classifier not being explicitly trained on these labels.

We emphasize that this generalization is not equivalent to fully open-vocabulary compositional reasoning. The strongest performance is still expected when the requested relation is close to the supervised primitive set or can be decomposed into these primitives. Highly abstract, ambiguous, or rare relations may require additional supervision or a richer relation vocabulary. Nevertheless, our results suggest that combining subject-token cross-attention with a strong frozen VLM prior allows the classifier to generalize beyond the exact training labels and provide effective inference-time guidance for novel compositional configurations.

\begin{table*}[h!]
\centering
\caption{\textbf{VBench evaluation across all dimensions for Wan2.2-14B.} Comparison of the base model with TTOM and \ourmethod{}.}
\label{tab:vbench_wan22_14b_full}
\setlength{\tabcolsep}{5pt}
\begin{tabular*}{\textwidth}{@{\extracolsep{\fill}} l c c c}
\toprule
\textbf{Dimension} &
\textbf{Wan2.2-14B} &
\textbf{+ TTOM} &
\textbf{+ \ourmethod{}} \\
\midrule
Subject Consistency        & 88.1\% & 89.0\% & \textbf{89.9\%} \\
Background Consistency     & 86.0\% & \textbf{86.8\%} & 86.6\% \\
Temporal Flickering        & 99.0\% & 99.1\% & \textbf{99.3\%} \\
Motion Smoothness          & 97.5\% & 98.1\% & \textbf{98.9\%} \\
Dynamic Degree             & 61.2\% & 63.5\% & \textbf{68.0\%} \\
Aesthetic Quality          & 66.2\% & 67.0\% & \textbf{67.2\%} \\
Imaging Quality            & 68.7\% & \textbf{69.1\%} & 68.9\% \\
Object Class               & 84.6\% & 85.4\% & \textbf{87.2\%} \\
Multiple Objects           & 62.7\% & 64.1\% & \textbf{67.5\%} \\
Human Action               & 80.8\% & 82.4\% & \textbf{85.7\%} \\
Color                      & 85.2\% & 85.8\% & \textbf{86.7\%} \\
Spatial Relationship       & 70.3\% & 74.6\% & \textbf{82.5\%} \\
Scene                      & 48.6\% & 49.3\% & \textbf{50.5\%} \\
Appearance Style           & 27.8\% & \textbf{28.6\%} & 28.2\% \\
Temporal Style             & 22.1\% & 22.7\% & \textbf{23.5\%} \\
Overall Consistency        & 27.6\% & 28.3\% & \textbf{29.1\%} \\
\midrule
Semantic Score             & 82.6\% & 84.6\% & \textbf{89.2\%} \\
Quality Score              & 84.6\% & 85.4\% & \textbf{85.6\%} \\
Final Score                & 83.6\% & 85.0\% & \textbf{86.3\% \textcolor{green!70!black}{(+2.7\%)}} \\
\bottomrule
\end{tabular*}
\end{table*}

\section{Details of Inference-Time Multi-Composition Guidance}
\label{app:multi_composition_guidance}

Beyond single-composition prompts, \ourmethod{} can naturally be extended to prompts that describe multiple compositional constraints over time. Many user instructions are temporally structured, such as ``a dog turns left and then turns right,'' where different composition constraints should hold at different parts of the video. In this setting, the goal is not to enforce a single global composition label over the entire generated video, but rather to guide different temporal segments toward different target compositions.

We formulate this setting by assuming that, in addition to the text prompt $p$, the user provides a set of temporally localized composition constraints:
\begin{equation}
\mathcal{Q}
=
\left\{
(y_i, \tau_i^{\mathrm{start}}, \tau_i^{\mathrm{end}})
\right\}_{i=1}^{M},
\end{equation}
where $y_i \in \mathcal{Y}$ is the target composition label for the $i$-th constraint, and
$\tau_i^{\mathrm{start}}$ and $\tau_i^{\mathrm{end}}$ denote the start and end times, in seconds, during which this composition should be expressed. For example, for the prompt ``a dog turns left and then turns right,'' the user may specify
\[
(y_1=\textit{left}, 0\text{s}, 3\text{s}),
\qquad
(y_2=\textit{right}, 3\text{s}, 5\text{s}).
\]
These temporal intervals are converted into frame ranges according to the video frame rate. Let $\Omega_i \subseteq \{1,\dots,F\}$ denote the set of frames corresponding to the temporal interval of composition label $y_i$.

At inference time, \ourmethod{} proceeds similarly to the single-composition setting. At a selected denoising step $t$, we extract the subject-token cross-attention maps and compute the aggregated representation $\phi(z_t,p)$. However, instead of applying the classifier to the full temporal representation, we restrict the representation to the frames associated with each temporal constraint. We denote this temporal masking operation by
\begin{equation}
\phi_i(z_t,p)
=
\mathrm{Mask}_{\Omega_i}\!\left(\phi(z_t,p)\right),
\end{equation}
where $\mathrm{Mask}_{\Omega_i}$ selects only the spatiotemporal features corresponding to the frames in $\Omega_i$.

The classifier is then applied separately to each temporal segment:
\begin{equation}
s_{t,i} = C(\phi_i(z_t,p)).
\end{equation}
For each composition constraint, we compute a cross-entropy loss with respect to the corresponding target composition label:
\begin{equation}
\mathcal{L}_{\mathrm{comp}}^{(i)}(z_t,p,y_i)
=
-\log s_{t,i}[y_i].
\end{equation}
The full multi-composition guidance objective is the sum of the segment-specific composition losses:
\begin{equation}
\mathcal{L}_{\mathrm{multi}}(z_t,p,\mathcal{Q})
=
\sum_{i=1}^{M}
\lambda_i
\mathcal{L}_{\mathrm{comp}}^{(i)}(z_t,p,y_i),
\end{equation}
where $\lambda_i$ controls the strength of each temporal composition constraint.

The latent is then updated by backpropagating the multi-composition loss:
\begin{equation}
z_t'
=
z_t
-
\eta_t
\nabla_{z_t}
\mathcal{L}_{\mathrm{multi}}(z_t,p,\mathcal{Q}).
\end{equation}

This formulation allows \ourmethod{} to enforce different compositional constraints at different temporal regions while keeping the underlying generation process unchanged. The same frozen text-to-video model, classifier, and subject-token cross-attention representation are used as in the single-composition setting. The only difference is that the classifier loss is computed on temporally masked portions of the latent video, so each composition label supervises only the frames in which it is intended to occur.

In practice, we apply multi-composition guidance at the same guided denoising steps used for the single-composition setting, as described in Sec.~\ref{sec:inference_time_control}. For each guided denoising step, all composition constraints are independently evaluated on their corresponding temporal segments, and their losses are accumulated before updating the latent. This enables temporally localized compositional control without training a separate model for multi-composition prompts. For example, given the prompt ``a dog turns left and then turns right,'' \ourmethod{} optimizes the frames corresponding to $0$--$3$ seconds toward the composition label \textit{left}, while optimizing the frames corresponding to $3$--$5$ seconds toward the composition label \textit{right}.

\begin{table*}[h!]
\centering
\caption{\textbf{VBench-Long evaluation across all dimensions for Rolling Forcing.} Comparison of Rolling Forcing with TTOM and \ourmethod{}.}
\label{tab:vbench_long_rolling_forcing_worldcontrol}
\setlength{\tabcolsep}{8pt}
\begin{tabular*}{\textwidth}{@{\extracolsep{\fill}} l c c c}
\toprule
\textbf{Dimension} &
\textbf{Rolling Forcing} &
\textbf{+ TTOM} &
\textbf{+ \ourmethod{}} \\
\midrule
Subject Consistency        & 86.02\% & 86.74\% & \textbf{87.31\%} \\
Background Consistency     & 85.41\% & 85.63\% & \textbf{85.86\%} \\
Temporal Flickering        & 82.49\% & 82.96\% & \textbf{83.42\%} \\
Motion Smoothness          & 90.50\% & 91.36\% & \textbf{92.18\%} \\
Dynamic Degree             & 51.24\% & 53.41\% & \textbf{56.93\%} \\
Aesthetic Quality          & \textbf{61.53\%} & 61.39\% & 61.21\% \\
Imaging Quality            & 62.67\% & \textbf{63.21\%} & 62.94\% \\
Object Class               & 82.14\% & 82.57\% & \textbf{83.08\%} \\
Multiple Objects           & 64.76\% & 64.89\% & \textbf{65.21\%} \\
Human Action               & 72.22\% & 74.05\% & \textbf{76.84\%} \\
Color                      & 83.08\% & 83.17\% & \textbf{83.29\%} \\
Spatial Relationship       & 75.42\% & 78.46\% & \textbf{81.37\%} \\
Scene                      & 27.51\% & \textbf{27.64\%} & 27.18\% \\
Appearance Style           & 19.35\% & 19.46\% & \textbf{19.62\%} \\
Temporal Style             & 19.52\% & 19.76\% & \textbf{20.08\%} \\
Overall Consistency        & 21.15\% & 21.39\% & \textbf{21.73\%} \\
\midrule
Semantic Score             & 59.13\% & 61.24\% & \textbf{63.82\%} \\
Quality Score              & 74.37\% & 74.92\% & \textbf{75.41\%} \\
Final Score                & 66.75\% & 68.08\% & \textbf{69.62\% \textcolor{green!70!black}{(+2.87\%)}} \\
\bottomrule
\end{tabular*}
\end{table*}

\section{Details of Autoregressive Multi-Composition Guidance}
\label{app:ar_multi_composition_guidance}

\ourmethod{} can also be applied to autoregressive long-video generation, where the video is generated as a sequence of temporal batches. This setting naturally supports prompts that describe different compositions at different stages of the video, such as ``a dog walks to the left of a bench, then moves behind it, and then stops in front of it.'' Instead of enforcing a single composition label over the entire video, we inject a new target composition at each generated batch.

Let the long video be generated as a sequence of batches
\begin{equation}
x = \{x^{(1)}, x^{(2)}, \dots, x^{(M)}\},
\end{equation}
where $x^{(m)}$ denotes the $m$-th generated batch. For each batch, we define a corresponding composition constraint
\begin{equation}
\mathcal{Q}_{\mathrm{AR}}
=
\left\{
(p_m, y_m)
\right\}_{m=1}^{M},
\end{equation}
where $p_m$ is the prompt segment used for the $m$-th batch, and $y_m \in \mathcal{Y}$ is the target composition label to be enforced in that batch. For example, the prompt ``a dog walks to the left of a bench, then moves behind it, and then stops in front of it'' can be decomposed into three batches with target composition labels \textit{left}, \textit{behind}, and \textit{front}, respectively.

At autoregressive batch $m$, the generator conditions on the previously generated context $x^{(<m)}$ and denoises the latent variable $z_t^{(m)}$ for the current batch. During the guided denoising steps, we extract the subject-token cross-attention maps for the current prompt segment $p_m$, compute the compositional representation $\phi(z_t^{(m)},p_m)$, and evaluate the classifier:
\begin{equation}
s_t^{(m)} = C(\phi(z_t^{(m)},p_m)).
\end{equation}
We then define a batch-specific composition loss:
\begin{equation}
\mathcal{L}_{\mathrm{AR}}^{(m)}(z_t^{(m)},p_m,y_m)
=
-\log s_t^{(m)}[y_m].
\end{equation}
The current batch latent is updated by backpropagating this loss:
\begin{equation}
z_t^{(m)\prime}
=
z_t^{(m)}
-
\eta_t
\nabla_{z_t^{(m)}}
\mathcal{L}_{\mathrm{AR}}^{(m)}(z_t^{(m)},p_m,y_m).
\end{equation}

This formulation allows the compositional objective to evolve over time while preserving temporal continuity through the autoregressive context. Previous batches provide visual and semantic history, while the classifier guides only the currently generated batch toward its assigned composition. Since the target composition is injected separately for each batch, \ourmethod{} can enforce temporally ordered compositional instructions without retraining the generator or requiring additional spatial controls.

In practice, we apply autoregressive multi-composition guidance at the same guided denoising steps used for the single-composition setting, as described in Sec.~\ref{sec:inference_time_control}. Thus, \ourmethod{} extends naturally to long-horizon autoregressive generation by changing the compositional objective at the batch level.

\section{Additional Experimental Implementation Details}
\label{app:implementation_details}
\paragraph{\textbf{Training data.}} We train the compositional classifier on a mixture of real and synthetic video data. For real data, we use VidOR and ImageNet-VidVRD ~\cite{shang2017video}, which provide supervision for object interactions and temporal compositions. For synthetic data, we use the dataset introduced in VChain, which increases the coverage of fine-grained compositional labels.

\paragraph{\textbf{Cross-attention extraction.}} For each video, we extract subject-token cross-attention maps from the frozen text-to-video model. For real videos, we first invert each video into the latent space of the frozen generator using the video inversion method of Yesiltepe et al.~\cite{yesiltepe2025dynamic}, and then replay the denoising trajectory to obtain the corresponding attention maps. We extract attention maps every 5 denoising steps, including the final clean step. The same set of denoising steps is used later during inference-time guidance.

\paragraph{\textbf{Composition leakage.}} To reduce composition leakage, we adopt a dual-inversion strategy~\cite{yiflach2025data}. For each training video, we invert the same sample twice: once with a prompt containing the correct composition label and once with a prompt containing an incorrect composition label sampled from the label vocabulary. Both inversions are supervised with the same ground-truth composition label. This prevents the classifier from relying on textual traces of the target label in the attention maps, and encourages it to infer the composition from the spatiotemporal evolution of the subject-token grounding.

\paragraph{\textbf{Classifier architecture.}} The compositional classifier consists of a learned aggregation module, a frozen Qwen2.5-VL-7B~\cite{qwen25vl} backbone, and a trainable linear classification head. The aggregation module compresses cross-attention maps extracted across layers, heads, and denoising steps into a compact spatiotemporal representation. This representation is passed to the frozen VLM backbone, followed by the classification head, which predicts the target composition label. During training, the text-to-video generator and the VLM backbone remain frozen; only the aggregation module and classification head are optimized.

\paragraph{\textbf{Guidance schedule.}} At inference time, the pretrained generator remains frozen and guidance is applied by optimizing the latent video variable using the compositional-classifier loss. We apply guidance only during the first 8 denoising steps, where coarse composition, layout, and motion structure are formed ~\cite{shaulov2025flowmo}. The later denoising steps are left unchanged, allowing the pretrained generator to refine appearance and fine visual details naturally.

\section{Additional Qualitative Results}
\label{sec:additional_qualitative}

\begin{figure*}[p]
  \centering
  \includegraphics[width=0.9\textwidth]{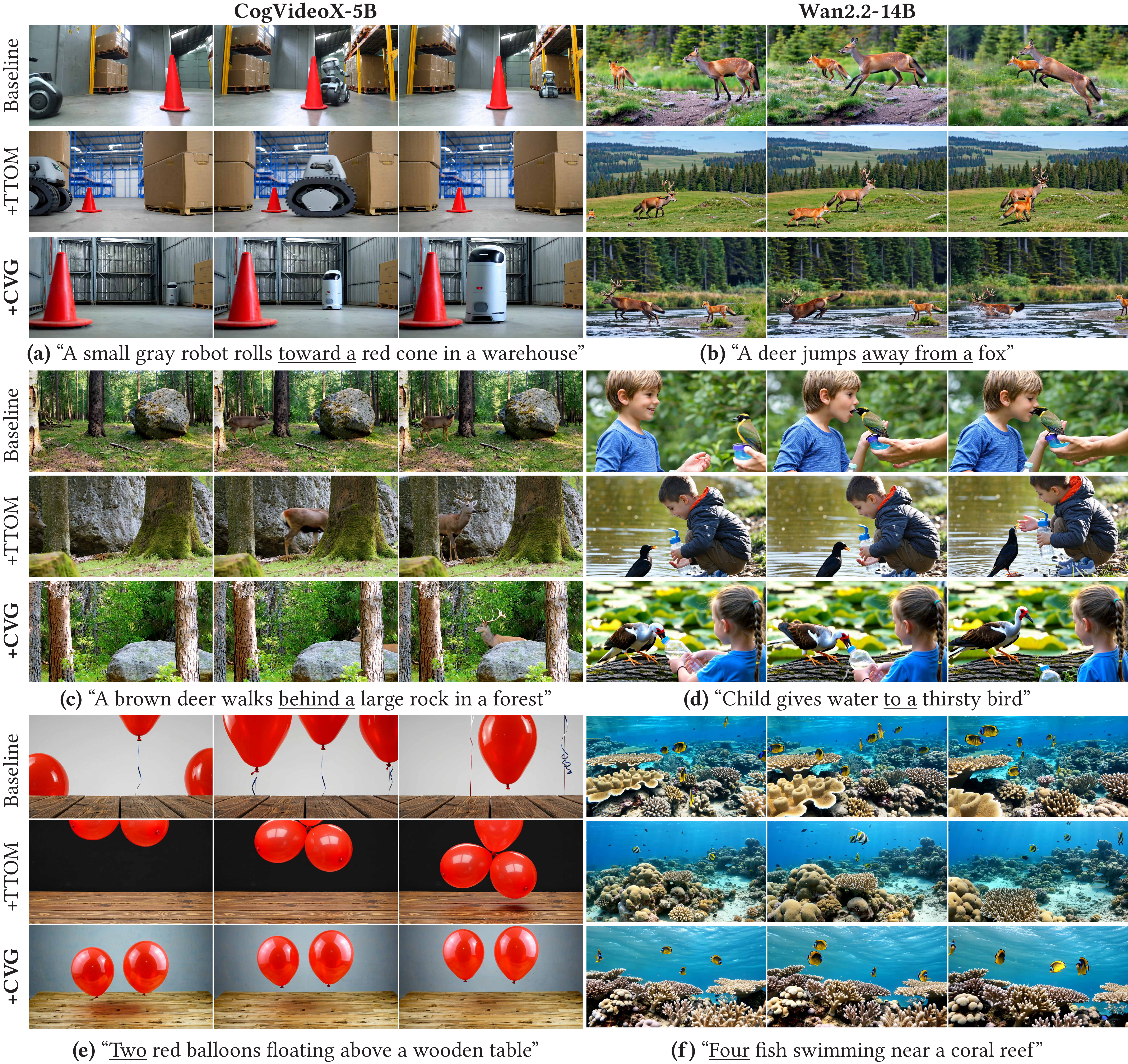}
  \caption{Additional Qualitative Results comparison between CVG and Wan2.2-14B (right), and between CVG and CogVideoX-5B (left).}
  \label{fig:additional_qualitative_res}
\end{figure*}

We provide further qualitative comparisons in Fig.~\ref{fig:additional_qualitative_res}, extending the analysis of Sec.~\ref{sec:qualitative_res} to additional compositional categories. As before, we compare CVG against the corresponding base generators, CogVideoX-5B~\cite{yang2024cogvideox} and Wan2.2-14B~\cite{wang2025wan}, and the inference-time optimization baseline TTOM~\cite{qu2025ttom}, using identical prompts and generation settings.

These examples reinforce the failure modes identified in the main paper, across a broader set of compositional categories. For \textbf{directional motion}, in Fig.~\ref{fig:additional_qualitative_res}(a), CogVideoX-5B and TTOM fail to roll the robot \emph{toward} the cone, leaving the two objects spatially disconnected; CVG instead produces a clear approach trajectory. In Fig.~\ref{fig:additional_qualitative_res}(b), Wan2.2-14B and TTOM place the deer alongside the fox rather than jumping \emph{away from} it, whereas CVG produces a clean separation between the two animals. For \textbf{spatial relations}, in Fig.~\ref{fig:additional_qualitative_res}(c), the baselines render the deer beside or in front of the rock; CVG correctly occludes the deer \emph{behind} the rock. For \textbf{action and interaction binding}, in Fig.~\ref{fig:additional_qualitative_res}(d), CogVideoX-5B and TTOM both mishandle the \emph{gives water to} interaction between the child and the bird, while CVG faithfully renders the directed action. For \textbf{numeracy}, in Fig.~\ref{fig:additional_qualitative_res}(e), CogVideoX-5B and TTOM generate three balloons instead of \emph{two}, and in Fig.~\ref{fig:additional_qualitative_res}(f), Wan2.2-14B yields an inconsistent number of fish across frames rather than \emph{four}; in both cases, CVG preserves the correct count throughout the rollout.

Together, these results confirm that CVG's compositional improvements generalize across a broader range of prompts beyond those shown in the main paper, without modifying the underlying generator.

\section{Ablation Study}
\label{sec:ablation}

We ablate the main design choices of \ourmethod{} on Wan2.2-14B using the average T2V-CompBench score across all compositional dimensions. As shown in Tab.~\ref{tab:ablation_avg_compbench_wan14b}, the full method achieves the best performance, reaching an average score of 0.594. Restricting guidance to only the first two denoising steps reduces the score to 0.583, suggesting that an overly short guidance window is insufficient to consistently establish the desired composition. Conversely, applying guidance at all denoising steps further decreases performance to 0.575, indicating that late-step optimization can over-constrain the generation and interfere with appearance refinement. Finally, removing leakage handling results in the lowest score, 0.568, highlighting the importance of the dual-inversion training procedure for preventing the classifier from relying on textual shortcuts rather than visual spatiotemporal grounding. Overall, these results support our design choice of applying composition guidance only during the early denoising steps while using leakage-aware classifier training.

\begin{table}[h!]
\centering
\caption{\textbf{Ablation study on composition-aware evaluation.} We report the average T2V-CompBench score for Wan2.2-14B across all compositional dimensions.}
\label{tab:ablation_avg_compbench_wan14b}
\setlength{\tabcolsep}{10pt}
\begin{tabular}{l c}
\toprule
\textbf{Method} & \textbf{Avg} \\
\midrule
\ourmethod{}                            & \textbf{0.594} \\
\ourmethod{} first 2 steps only         & 0.583 \\
\ourmethod{} all denoising steps        & 0.475 \\
\ourmethod{} w/o leakage handling       & 0.568 \\
\bottomrule
\end{tabular}
\end{table}

\end{document}